# The non-algorithmic side of the mind

**Paola Zizzi**

Department of Psychology, University of Pavia,
Piazza Botta, 6, 27100 Pavia, Italy
paola.zizzi@unipv.it

**Abstract**
The existence of a non-algorithmic side of the mind, conjectured by Penrose on the basis of Gödel's first incompleteness theorem, is investigated here in terms of a quantum metalanguage.
We suggest that, besides human ordinary thought, which can be formalized in a computable, logical language, there is another important kind of human thought, which is Turing-non-computable. This is methatought, the process of thinking about ordinary thought. Metathought can be formalized as a metalanguage, which speaks about and controls the logical language of ordinary thought. Ordinary thought has two computational modes, the quantum mode and the classical mode, the latter deriving from decoherence of the former. In order to control the logical language of the quantum mode, one needs to introduce a quantum metalanguage, which in turn requires a quantum version of Tarski Convention T.

# 1. Introduction

There are three different ways by which fundamental high-level mental activities do manifest themselves [1]. Two ones are algorithmic (Turing-computable): the classical computational mode, and the quantum computational mode. The third one is non-computable.

The quantum mode concerns extremely fast mental processes of which humans are mostly unaware of, and is logically described by the logic of quantum information and quantum computation, called Lq [2]. The atomic propositions of Lq are interpreted as the basis states of a complex Hilbert space, while the compound propositions are interpreted as qubit states. Therefore, the physical model of the quantum mode of the mind is Quantum Information. The classical mode concerns those mental processes, which humans are aware of. It arises from the decoherence of the quantum computational state, and is logically described by a sub-structural, non-classical logic called Basic Logic (BL) [3]. In a sense, the quantum mode "prepares" the classical mode, which otherwise would take very long to perform even the easiest tasks, but most of the quantum information remains hidden. The classical mode comes in to play by "flashes" of decoherence, which occur so often that humans get the impression of a "flow of consciousness".

The third mode, which is non-algorithmic, concerns metathought (intuition, intention and control) and is described by a quantum metalanguage (QML) [2], which "controls" the logic Lq of the quantum mode. The assertions of the QML are physically interpreted as the field states of a dissipative quantum field theory (DQFT) of the brain [4] [5].

The atomic propositions of the quantum object-language (QOL) [2] are asserted, in the quantum metalanguage (QML) with an assertion degree, which is a complex number. We show that this fact requires that the atomic propositions in the QOL are endowed with a fuzzy modality "Probably" [6] and have fuzzy (partial) truth-values [7], which sum up to one.

The QML is the language of metathought. The very importance of metathough, which deals with intuition, intention and control, resides in the fact that it distinguishes humans from machines. In fact, the language of metathough, which is non-algorithmic, being described by a metalanguage, cannot be acquired independently by a machine, which is endowed only with an object-language.

The physical interpretation of this impossibility, is the irreversibility of the reduction process [8] from the DQFT of the brain to the quantum-computational theory of the mind.

The question of machine implementation of (human-like) mental processing is very old and dates to the early days of Artificial Intelligence. As well known, in 1950 A. M. Turing [9] adopted a purely behavioural criterion (instantiated through his famous test) to detect the occurrence of mental processes in a machine. Within this approach, a machine was recognized as endowed with a mind when its behaviour was indistinguishable from the one of a human being performing (supposed) mental operations. Later this attitude gained a wide popularity when Cognitive Psychology adopted the Computational Symbolic Approach [10] speaking of the (functional) equivalence between mind and a digital computer. In the Eighties the philosophical considerations already made by Searle [11] and others, began to cast serious doubts about the validity of this definition of mind.

The need of a logic for (quantum) reasoning raises from realizing the fact that humans have basic logical rules but also some fundamental metalogical ones. Moreover, there are very few structural rules. In other words, the logic of reasoning is much more concrete and weaker than other abstract and structural logics, like for example Aristotelian classical logic. This requires a sub-structural logic, which can be viewed as the general platform for any other logic. All these requirement were met in BL [3] in the classical case (or classical mode). A quantum version of BL, called Lq, was introduced in [2]. In Lq, two new logical connectives were introduced, the connective "quantum superposition" (the quantum version of the classical conjunction) and the connective "entanglement". Finally, the probabilistic feature of any quantum theory is also present in Lq, because the partial truth values, whose range is the real interval [0,1] are interpreted as probabilities. This takes into account the fuzzy and probabilistic features of some non-formalized aspects of (non-ordinary) thought.

In this context, we shall try to clarify Penrose's conjecture [12] on the non-computational aspects of the mind in relation with Gödel's First Incompleteness Theorem.[13]. Penrose claims that a mathematician can assert the truth of a Gödel sentence *G*, although the latter cannot be demonstrated within the axiomatic system, because he is capable of recognizing an indemonstrable truth due to the non-algorithmic aspect of the mind. In our opinion, the fact that the mathematician can assert the truth of *G*, is that he is using the non-computable mode of metathought described by the metalanguage, where assertions stand, and where Tarski introduced the truth predicate [14]. Furthermore, the fuzzy-probabilistic features of QML, induce to modify Tarski Convention *T* as Convention *PT* (where *P* stands for "*Probably*").

This paper is organized as follows.

In Sect. 2, we give a brief review of the basic concepts of classical and quantum metalanguages, and the respective definitional equations.

In Sect. 3, we show that the reflection principle of Basic Logic can be reformulated in terms of Tarski Convention T and T-Schema.

In Sect. 4, we reformulate the reflection principle in the quantum case (the logic Lq) in terms of a modified version of Convention T, namely Convention PT (where P stands for "Probably"), and T-Schema.

In Sect. 5, we apply Convention PT to the Gödel sentence.

Sect. 6 is devoted to the conclusions.

## 2. Classical and quantum metalanguages: A brief review

A metalanguage (ML) is a language which talks about another language, called object-language (OL).

A (classical) formal ML consists of (classical) assertions, and meta-linguistic links among them. (By classical assertions, we mean assertions which are stated with certitude). It consists of:

i) Atomic assertions: $\vdash A$ (*A* declared, or asserted), where A is a proposition of the OL.

ii) Meta-linguistic links: $\vdash$ ("yields", or "entails"), and (metalinguistic "and").

iii) Compound assertions. Example: $\vdash A$ and $\vdash B$.

Let us consider the introduction of the logical connective & in Basic Logic (BL) [3].

In the OL, let A, B be propositions.

In the ML, I read: A decl. , B decl, that is: $\vdash A$ , $\vdash B$ respectively (where "decl." is the abbreviation of "declared", which also can mean "asserted").

Let us introduce a new proposition A&B in the OL. In the ML, we will read: A&B decl., that is: $\vdash A \& B$.

The question is: From A &B decl., can we understand A decl. and B decl. ?

More formally, from $\vdash A \& B$ can we understand $\vdash A$ and $\vdash B$ ?

To be able to understand A decl. and B decl. from A&B decl, we should solve:

$$\vdash A \& B \quad \underline{iff} \quad \vdash A \quad \underline{and} \quad \vdash B \tag{2.1}$$

where "*iff*" stands for "*if and only if*".

Eq. (2.1) is the definitional equation of the connective & in BL [3].

A quantum metalanguage (QML) [2] consists of:

i) Quantum atomic assertions: $\vdash^\lambda p$ \hfill (2.2)

where *p* is a proposition of the quantum object-language (QOL), and $\lambda$ is a complex number, called the assertion degree, which indicates the degree of certitude in stating the assertion. In the limit case $\lambda = 1$, quantum assertions reduces to classical ones. The truth-value of the corresponding proposition *A* in the QOL, is given by:

$$v(p) = |\lambda|^2 \in [0,1] \tag{2.3}$$

which is a partial truth-value as in Fuzzy Logic [7].
ii) Meta-linguistic links: $\vdash$ ("yields", or "entails"), and (metalinguistic "and"), as in the classical case.
iii) Compound assertions. Example: $\vdash^{\lambda_0} p_0$ and $\vdash^{\lambda_1} p_1$
iv) Meta-data:
$$\sum_{i=0}^{n-1} v(p_i) = 1 \qquad (2.4)$$
where $n$ is the number of atomic propositions in the QOL.
As in the classical case, one should solve the definitional equation of the quantum connective $_{\lambda_0}\&_{\lambda_1}$ [2]:
$$\vdash p_0\,{}_{\lambda_0}\&_{\lambda_1} p_1 \quad \textit{iff} \quad \vdash^{\lambda_0} p_0 \quad \text{and} \quad \vdash^{\lambda_1} p_1 \qquad (2.5)$$
with the constraint:
$$|\lambda_0|^2 + |\lambda_1|^2 = 1 \qquad (2.6)$$
which is the meta-data in Eq. (2.4) written in terms of the assertion degrees, by the use of Eq. (2.3).

## 3. The reflection principle of BL and Tarski "Convention T"
By Tarski Convention $T$ [14], every sentence $p$ of the object-language (OL) must satisfy:
*(T)*: '$p$' is true *iff* $p$     (3.1)
where '$p$' stands for the name of the proposition $p$, which is the translation in the metalanguage ML of the corresponding proposition in the OL.
The standard example is:
'*Snow is white*' is true *iff* snow is white.
Convention *T* is also called "*material adequacy condition*", in the sense that a sentence is true if it denotes the existing state of affairs (or, if it is conform to reality).
By the point of view of a physicist this would mean that a sentence is true if it states something that is observable, measurable, computable. Obviously, in the classical context of Tarski, a true sentence has truth value 1, which corresponds to probability 1 in the measurement procedure.
But this state of affairs changes when we deal with a quantum metalanguage, as we will see in the following.
Tarski *T*-Schema (*equivalence schema*) [14] allows to state inductively the truth of compound propositions.
For example, for the conjunction A&B of two propositions A and B, the *T*-Schema gives:
'A & B' is true *iff* A is true and B is true.     (3.2)
There is a close relation between the concepts of assertion and truth. Then, we will "translate" Tarski Convention *T* and *T*-Schema in terms of assertions and metalinguistic links to recover the definitional equation of the reflection principle. We do so for a precise scope, that is, to show that the mathematician asserting the truth of the Gödel sentence *G* in his (non-algorithmic) metalanguage is operating in Tarski semantic theory of truth, where the material adequacy condition holds.
As we will see, the "translation" mentioned above is quite easy in the case of a classical metalanguage, while, for the quantum case, one is led to borrow some concepts from probability and fuzzy logic.
Let us start with the classical case.
Let us apply Convention *T* to the two sentences A and B of the object-language:
*(T)*: 'A' true *iff* A     (3.3)
*(T)*: 'B' true *iff* B     (3.4)
The *T*-Schema gives:
'A & B' true *iff* 'A' true and 'B' true     (3.5)

In terms of assertions, we have:

$$\vdash 'A' \quad iff \quad A \tag{3.6}$$

$$\vdash 'B' \quad iff \quad B \tag{3.7}$$

From A and B in the OL, we can form the compound proposition $A \& B$, to which we apply again Convention *T*:

$$(T): \vdash '(A \& B)' \quad iff \quad A \& B \tag{3.8}$$

The *T*-Schema gives, for the LHS of Eq. (3.8):

$$\vdash '(A \& B)' \quad iff \quad \vdash 'A' \text{ and } \vdash 'B' \tag{3.9}$$

We are now allowed to discard the quotation marks, as they appear on both sides of Eq. (3.9), and we get:

$$\vdash A \& B \quad iff \quad \vdash A \text{ and } \vdash B \tag{3.10}$$

which is the (classical) definitional equation for the (classical) logical connective & in BL.

## 4. The Convention PT.

The quantum case is based on a different kind of Convention *T*, namely the Convention "*Probably*" *T*, which we will introduce in the following.

Before, we shall recall two notions of fuzzy logic and modal logic.

The fuzzy notion *probably* can be axiomatized as a *fuzzy modality*. Having a probability on Boolean formulas, define for each such formula $\varphi$ a new formula $P(\varphi)$, read "probably $\varphi$"[6] and define the truth value of $P(\varphi)$ to be the probability of $\varphi$, that is:

$$v(P\varphi) = p(\varphi) \in [0,1] \tag{4.1}$$

Following Hajek, we stress the fact that it was widely believed that truth values in fuzzy logic had to be totally distinguished from probabilities, as fuzzy logic [7] is understood truth-functional (by t-norms), and probability is not functional-preserving. However, there is a way to get a bridge between fuzziness and probability, that is, to take the probability of a Boolean formula of the classical propositional calculus to be the truth value of the fuzzy proposition: "$\varphi$ is probable" denoted by $P\varphi$ (this makes P to a fuzzy modality). Over the formulas $P\varphi$ as new atoms one builds a fuzzy logic, preferably Łukasiewicz logic [15]. The new logic is named *FP*(Ł) (Fuzzy Probability over Ł) [6].

Let us consider a set S of N atomic Boolean propositions of OL:

$$\psi_i \quad (i=1,2,......N)$$

Let us call $p_i$ $(i=1,2.......n)$ with $n < N$, the propositions of a subset $S' \subset S$, to which it is possible to assign a probability $p$ such that:

$$\sum_{i=1}^{n} p(p_i) = 1. \tag{4.2}$$

Then, we can define *n* new propositions $P(p_i)$, for which it holds:

$$v(P(p_i)) = p(p_i) \in [0,1]. \tag{4.3}$$

And, from Eq. (4.2) it follows:

$$\sum_{i=1}^{n} v(P(p_i)) = 1 \tag{4.4}$$

We can then reformulate Tarski Convention *T* for any sentences $P(p_i)$ as convention *PT*.

$$(PT): \text{'}p_i\text{'} \text{ is probably true } iff \quad P(p_i). \tag{4.5}$$

Example: The proposition '*Snow is white*" is probably true if and only if probably snow is white. The expression "*is probably true*" means that I am asserting the truth of a sentence with a certain degree of assertion, not with complete certitude.

In terms of assertions, convention *PT* reads:

$$\vdash^{\lambda_i} \; 'p_i' \quad \underline{iff} \quad P(p_i) \tag{4.6}$$

which means that proposition '$p_i$' is asserted with assertion degree $\lambda_i$ if and only if *probably* $p_i$, with probability $|\lambda_i|^2 \in [0,1]$.

From Eqs. (4.5) and (4.6) it follows that by assigning a probability to a sentence $p_i$ of a classical OL, the corresponding assertion belongs to a QML, that is, the fuzzy probabilistic proposition $P(p_i)$ does not belong anymore to the classical OL, but to a QOL.

The truth value of $P(p_i)$ is just the probability of $p_i$, which is the squared modulus of the assertion degree $\lambda_i$:

$$v(P(p_i)) = p(p_i) = |\lambda_i|^2 \tag{4.7}$$

Let us consider two probabilistic propositions $p_0, p_1$ of the OL.

For $p_0$ we get, from Convention *PT*:

$$\vdash^{\lambda_0} \; 'p_0' \quad \underline{iff} \quad P(p_0) \qquad \text{with: } v(P(p_0)) = p(p_0) = |\lambda_0|^2 \tag{4.8}$$

In the same way, we get, for $p_1$:

$$\vdash^{\lambda_1} \; 'p_1' \quad P(p_1) \qquad \text{with: } v(P(p_1)) = p(p_1) = |\lambda_1|^2 \tag{4.9}$$

Let us now form, in the QOL, the new conjunction $_{\lambda_0}\&_{\lambda_1}$ taking into account the weights $\lambda_0, \lambda_1$ by which the two propositions $p_0, p_1$ contribute to the conjunction itself.

We define then:

$$p_0 \;_{\lambda_0}\&_{\lambda_1} p_1 \equiv P(p_0)\,\&\,P(p_1). \tag{4.10}$$

We can then apply Convention *T* to the new formed proposition $P(p_0)\,\&\,P(p_1)$:

$$\vdash P(p_0)\,\&\,P(p_1) \quad \underline{iff} \quad P(p_0)\,\&\,P(p_1). \tag{4.11}$$

From Eqs. (4.10) and (4.11) it follows:

$$\vdash 'p_0 \;_{\lambda_0}\&_{\lambda_1} p_1' \quad \underline{iff} \quad p_0 \;_{\lambda_0}\&_{\lambda_1} p_1 \tag{4.12}$$

By applying the *T*-Schema to the LHS of Eq. (4.12) one gets:

$$\vdash p_0 \;_{\lambda_0}\&_{\lambda_1} p_1 \quad \underline{iff} \quad \vdash^{\lambda_0} p_0 \quad \underline{and} \quad \vdash^{\lambda_1} p_1 \tag{4.13}$$

which is the definitional equation for the connective $_{\lambda_0}\&_{\lambda_1}$ of quantum superposition [2].

## 5. The non-computational mode of the mind

Since Hilbert's program, all true mathematical statements were assumed to be provable within the formal axiomatic system. This assumption was shown to be wrong by Gödel's First Incompleteness Theorem and Turing's Halting Problem [9] for which there exist true statements which are not provable within the formal system.

Gödel's First Incompleteness Theorem states that: Any effectively generated formal system capable of expressing arithmetic, cannot be both consistent and complete.

Here "effectively generated" means that in principle there exist a computer program which can enumerate all the axioms of the system, "consistent" means that there is no statement of the system, such that both the statement and its negation are provable from the axioms, and "complete" means that for any statement of the system, either the statement or its negation are provable from the axioms.

In particular, for any effectively generated, consistent formal system *F* that includes arithmetic, there is a statement which is true, but not provable within the theory. Such a statement is called the Gödel sentence *G(F)*.

In this Section, we will review, in the light of the QML, Penrose's conjecture that some aspects of the mind have a non-algorithmic nature in relation with Gödel's First Incompleteness Theorem.

Penrose bases his conjecture on the fact that the human mind is able to recognize the truth of the Gödel sentence $G(F)$ although the latter is not demonstrable within the axiomatic system.

The Gödel sentence $G(F)$ is:

$G(F)$= "This sentence cannot be proved in $F$".

Penrose says that the First Incompleteness Theorem tells us that no computer, working within a consistent formal system $F$ can prove the sentence $G(F)$, while we humans can "see" the truth of $G(F)$. In fact, we "see" that $G(F)$ is true, because, if it were false, then it would be provable in $F$, which is absurd, because $G(F)$ states that it cannot be proved in $F$.

In the task of recognizing the truth of $G(F)$ the human mind can develop mathematical insight, or intuition, a property which is not shared by any algorithmically based system of logic. This seems quite reasonable to us. In fact, mathematical intuition is described by a quantum metalanguage, which is non-algorithmic. As a metalanguage cannot be given to a machine, which uses only the object-language, it is obvious that humans and machines have different levels of language. We humans have both the ML by which we give instruction to the machine, and the OL, already contained in the ML., while machines can utilize only the OL.

In particular, QML organizes and controls our own QOL. When the mathematician asserts the truth of $G(F)$, in fact he is operating at the level of QML, where assertions (with a degree of assertion) live.

In fact, what the mathematician asserts is:

$$\mid\!\!-^{\lambda} G(F) \tag{5.1}$$

As we have seen in the previous Section, this is equivalent to the probabilistic version of Tarski Convention $T$, namely to $PT$:

$(PT)$: '$G(F)$' is probably true iff $P(G(F))$ (5.2)

with:

$$v(P(G(F))) = p(G(F)) = |\lambda|^2 \tag{5.3}$$

By the Second Incompleteness Theorem, it holds:

$G(F) = Con\,(F)$ (5.4)

From Eqs. (5.2) and (5.4) it follows:

$P(G(F))=P(Con(F))$ (5.5)

That is, the probabilistic character of the truth of the Gödel sentence reflects into the probability that the formal system $F$ is consistent.

We humans can recognize the truth of $G(F)$ with a certain probability, but by doing so we affect in some way the consistency of the formal system. In other words, by saying that the Gödel sentence is *probably true*, we say that the formal system is not totally consistent, and consequently, is not totally incomplete, but only probabilistically incomplete.

## 6. Conclusions

The original conjecture of Penrose about the existence of non-algorithmic aspects of the mind regarded mainly consciousness. However, conscious, rational human thought consists of a very rapid sequence of decoherence processes from the quantum computational mode to the classical one. More specifically, superposed tubulins/qubits decohere to classical bits [16] at a fast rate.

In other words, consciousness is made of "flashes" of classical computation. Consciousness cannot be identified with the classical mode of the mind, because that would lead to an absurd conclusion: A classical Turing machine, which persists in the classical mode, would be more "conscious" than a human mind.

The problem is that the static conscious state of a classical computer is totally useless for any kind of aware reasoning, which is dynamical by definition. It is the never-ending supply of new data coming from decoherence, which makes the difference.

What is really non-algorithmic, is the origin of consciousness in the quantum metalanguage, not consciousness itself. Now, a question naturally arises: Which is the logical counterpart of decoherence? In [2], we found that the quantum cut-rule can be physically interpreted in terms of a projective quantum measurement. However, the (quantum) cut-rule is a meta-rule, and pertains to (quantum) metalanguage, not to the (quantum) object-language from which one would expect decoherence to the classical mode. This fact suggests that decoherence (and then consciousness) has its roots in the quantum metalanguage. Consequently, consciousness comes from a non-computational mode.

## Acknowledgements

I wish to thank Eliano Pessa for useful discussions.